\documentclass{article}

     \usepackage[preprint]{neurips_2023}

\usepackage[utf8]{inputenc} 
\usepackage[T1]{fontenc}    
\usepackage{hyperref}       
\usepackage{url}            
\usepackage{booktabs}       
\usepackage{amsfonts}       
\usepackage{nicefrac}       
\usepackage{microtype}      
\usepackage{xcolor}         

\usepackage{amssymb}
\usepackage{amsmath}
\usepackage{bm}
\usepackage{graphicx}
\usepackage[ruled,vlined]{algorithm2e}

\newcommand{\bz}{\mathbf{z}}

\newcommand{\vf}{f}

\newcommand{\vphi}{\phi}
\newcommand{\expect}{\mathbb{E}}
\newcommand{\eq}[1]{eq.(\ref{#1})}
\newcommand{\fig}[1]{Fig.(\ref{#1})}
\newcommand{\ouralgo}{{\em DBShap} }

\author{
Narayanan U. Edakunni \\
American Express\\
\And
Utkarsh Tekriwal \\
American Express\\
\And
Anukriti Jain\\
American Express
}

\title{Explaining Drift using Shapley Values}

\begin{document}

\maketitle

\begin{abstract}
Machine learning models often deteriorate in their performance when they are used to predict the outcomes over data on which they were not trained. These scenarios can often arise in real world when the distribution of data changes gradually or abruptly due to major events like a pandemic. There have been many attempts in machine learning research to come up with techniques that are resilient to such Concept drifts. However, there is no principled framework to identify the drivers behind the drift in model performance. In this paper, we propose a novel framework - \ouralgo that uses Shapley values to identify the main contributors of the drift and quantify their respective contributions. The proposed framework not only quantifies the importance of individual features in driving the drift but also includes the change in the underlying relation between the input and output as a possible driver. The explanation provided by \ouralgo can be used to understand the root cause behind the drift and use it to make the model resilient to the drift.
\end{abstract}


\section{Introduction}
Proliferation of complex machine learning models that model data at scale, has led to challenges in maintaining these models and keeping it current with respect to the changes in the environment. The characteristics of data on which a model operates, often change over time. The change could be gradual (eg. change in purchase power due to inflation) or abrupt (eg. change in spend patterns due to Covid). Over the years, there has been a lot of research dedicated to the study of such changes and their effect on the performance of machine learnt models. These changes have often been termed as concept drift, model drift or data drift. Different aspects of concept drift have been explored in detail by \cite{survey}. The survey concludes that while there has been a lot of research dedicated to identifying drift (\cite{detect1,gama,detect2}) and adapting to drift (\cite{adapt1,adapt2}), not much work has been done on the problem of identifying the root cause of drift. In this paper, {\em we propose a novel methodology that uses Shapley values to identify the cause of model drift and measure the contribution of different components of the data to overall deterioration of model predictions}.

In many applications, we are interested in understanding the effect of distributional change in the data affecting the quality of predictions made by a machine learnt model. For instance, in case of a credit risk model, we observe a change in the distribution of risk score across the population after a recession or a pandemic. We would then like to understand the reason behind this shift. Specifically, we would like to understand if there are one or more inputs whose distribution had changed, leading to the change in the risk score or has the underlying relation between the input and output itself changed. We often term these changes as {\em drift}, with the drift caused due to changes in the distribution of the input termed as {\em temporary drift}(\cite{laz}), {\em virtual drift}(\cite{gama}), {\em sampling shift}(\cite{sal}) and {\em feature change}(\cite{gao}).  Change in the functional relation between input and output has been variously referred to as {\em real drift}(\cite{gama}),{\em concept shift}(\cite{sal}) and {\em conditional change}(\cite{gao}). In this paper, we adopt the term {\em real drift} to refer to the drift in the relation between the input and the output. We refer to a change in the distribution of the input data as {\em virtual drift}. Conventionally, the term concept drift is used to refer to the change in characteristics of data over time. In this paper we expand the definition of drift from being a change over time to any change in the characteristics of the data over two different subpopulations. For example, the risk of diabetes predicted by a model for individuals living in a certain geographical area might be different from another geographical area. In this case, we would like to understand the factor contributing to the difference in risk of diabetes separated across a spatial area rather than across time. In another example, we might want to compare the predictions of a model built for crime recidivism over different race categories. It is important to note that in all these examples our aim is to identify the factors contributing to the difference in the behaviour of the model over different populations. One of the important contributions of the paper is that the the list of factors contributing to the change, not only includes the change in the distribution of individual features(virtual drift) but also includes the change in the functional relation between input and output (real drift). We achieve this distinction between real and virtual drift by a novel adaptation of Shapley value(\cite{shap}) to operate on distributions instead of scalar valued features.

Shapley value has been reframed into a tool to measure the importances of feature in making a prediction by a model, given the test input. In recent times, there have been many definitions and variants of Shapley values to measure the importance of features used in a model. Out of these definitions, the most useful is BShap(\cite{bshap}) that starts with the axiom that importances are always relative and is measured relative to a reference point. Hence, baseline Shapley measures the contribution of a change in the values of features between target and reference points that led to the difference in the output of a function. In this paper, we utilise this property of baseline Shapley to measure and quantify the relative contributions of changes in feature distributions in affecting the output of a model.
\section{Related work}
Concept drift has been well studied in terms of detecting and alleviating the effects of concept drift on machine learning models. In these studies, the focus has been either in detecting the drift(\cite{detect1,gama,detect2}) or adapting to the drift(\cite{adapt1,adapt2}). The only study that looks at the use of explainability to understand the drivers behind concept drift is \cite{processmining} which develops a framework for explaining drift in the context of process mining. The solution that is presented in the work is specific to process mining and can be applied only for drift across time. In our paper, we develop the framework for explanation using a more principled framework of Shapley values and hence is also generic in its scope and can be applied across different situations that involve drift.

In this paper, we first come up with a novel framework to compute Shapley values for functions with probability distributions as inputs. Using this framework, we can track changes in data distribution and the effect of these changes on the distribution of the output variable. We can then combine this form of Shapley values designed for distributions, along with baseline Shapley, to obtain the feature importances of changes that led to the change in the distribution of the output. This can then be used to understand the root cause behind the drift.
\section{Shapley values for function over distributions}
\label{sec:dist}
In this section, we derive a mathematical framework of Shapley values that can be used when dealing with functions over distributions and in turn to derive the importance of change in distributions. Let $z_1,z_2 \hdots z_n$ denote random variables and $P(z_1 \hdots z_n)$ a distribution over them. For convenience, we assume that $z_1,z_2 \hdots z_n$ are all discrete valued such that $z_1,z_2 \hdots z_n$ take values from the sets $Z_1,Z_2 \hdots Z_n$ respectively.  We can now define a generic vector valued function $\vf$ over the distribution of these variables denoted by - 
\begin{equation}
\vf(P(z_1,z_2 \hdots z_{m_1}),P(z_{m_1+1} | z_{m_1 + 2} \hdots z_{m_2}) \hdots P(z_{m_j},z_{m_j + 1} \hdots z_n))
\label{eqn:vecfn}
\end{equation}
The specific factorization of the joint distribution $P(z_1 \hdots z_n)$ into $P(z_1,z_2 \hdots z_{m_1}),P(z_{m_1+1} | z_{m_1 + 2} \hdots z_{m_2}) \hdots P(z_{m_j},z_{m_j+1} \hdots z_n)$ as given in \eq{eqn:vecfn} is one possibility, but in general the factorization can be arbitrary and is dictated by the function $\vf$ and the dependence structure amongst the variables. 
We can now compare the values of $f$ for two different distributions- we denote one by $P_b(\bz)$ for the baseline distribution and the other by $P_t(\bz)$ for the target distribution, where $\bz = \{z_1,z_2 \hdots z_n\}$. The corresponding values for $\vf$ would be $\vf(P_{b}(z_1,z_2 \hdots z_{m_1}),P_{b}(z_{m_1+1} | z_{m_1 + 2} \hdots z_{m_2}) \hdots P_{b}(z_{m_j},z_{m_j+1} \hdots z_n))$ and $\vf(P_{t}(z_1,z_2 \hdots z_{m_1}),P_{t}(z_{m_1+1} | z_{m_1 + 2} \hdots z_{m_2}) \hdots P_{t}(z_{m_j},z_{m_j+1} \hdots z_n))$ respectively.

We can simplify the notation and analysis, by introducing surrogate features $s_1,s_2 \hdots s_k$ which takes values from $\{0,1\}$. These surrogate variables are defined based on whether we want to model a baseline probability distribution or a target distribution within the function $\vf$. We can define a new function $\vf'$ over $s_1,s_2 \hdots s_k$, equivalent to $\vf$ as -
\begin{eqnarray}
&&\vf'(s_1,s_2 \hdots s_k) = \vf\left(I[s_1 = 0]P_{b}(z_1,z_2 \hdots z_{m_1}) + I[s_1 = 1]P_{t}(z_1,z_2 \hdots z_{m_1}), \right. \nonumber \\
&& \left. I[s_2 = 0]P_{b}(z_{m_1+1} | z_{m_1 + 2} \hdots z_{m_2}) + I[s2 = 1]P_{t}(z_{m_1+1} | z_{m_1 + 2} \hdots z_{m_2}) \hdots\right)
\label{eqn:s}
\end{eqnarray}
where $I[.]$ is an indicator function that takes value $1$ when its argument evaluates to true else takes a value of $0$. Therefore, we have the following correspondence between $\vf'$ and $\vf$ -
\begin{eqnarray}
\vf'(s_1=0,s_2=0 \hdots s_k=0) &=& \vf(P_{b}(z_1,z_2 \hdots z_{m_1}),P_{b}(z_{m_1+1} | z_{m_1 + 2} \hdots z_{m_2}) \hdots \nonumber \\
&&P_{b}(z_{m_j},z_{m_j+1} \hdots z_n)) \\
\vf'(s_1=1,s_2=1 \hdots s_k=1) &=& \vf(P_{t}(z_1,z_2 \hdots z_{m_1}),P_{t}(z_{m_1+1} | z_{m_1 + 2} \hdots z_{m_2}) \hdots \nonumber \\
&&P_{t}(z_{m_j},z_{m_j+1} \hdots z_n))
\end{eqnarray}
which means that when all of the values of $s$ is zero, $f$ outputs the baseline value and when all the values of $s$ is $1$ it outputs the target value. For all other combinations of $0$ and $1$ values of $s$ value of the function is computed according to \eq{eqn:s}.

We can now derive the expression for feature importances for a function of the form of $\vf'$. We start with the definition of Shapley importances for generic functions -
\begin{equation}
\phi(s_i) = \sum_{S \subset N \setminus \{s_i\}}{\frac{|S|!(n-|S|-1)!}{n!}}\left(v(S \cup \{s_i\}) - v(S)\right)
\label{eqn:shapvec}
\end{equation}
Here, $N = \{s_1 \hdots s_k\}$, $S$ is a subset of $N$ excluding the variable $s_i$. The quantity $v$ denotes the value function which provides the value of a particular coalition of features. 

In this paper, we are interested in the case where we want to compare two different distributions and understand the cause behind the change. We use Baseline Shapley(\cite{bshap}) to compute the Shapley values that measure the importances of change observed in the input values. The value function $v$ in our case would be $f'$ and the features would be the set of surrogate features $s_1 \hdots s_k$. Accordingly, we modify \eq{eqn:shapvec} to get -
\begin{equation}
\phi(s_i) = \sum_{S \subset N \setminus \{s_i\}, T = N \setminus \{S,s_i\}}{\frac{|S|!(n-|S|-1)!}{n!}}(\vf'(S^{t} \cup \{s_i^{t}\} \cup T^{b})
 - \vf'(S^{t} \cup \{s_i^{b}\} \cup T^{b})) \label{eqn:phivec}
\end{equation}
where the equation has been modified such that the variables that do not take part in the coalition takes the base value, indicated by superscript $b$. The variables that take part in the coalition take the value in the target data, indicated by superscript $t$. From \eq{eqn:s} we know that when a feature of $f'$ takes the value from baseline, the surrogate feature $s$ takes value $0$ and when it takes value from target, the value of $s$ is $1$. Hence, we can rewrite \eq{eqn:phivec} as -
\begin{eqnarray}
\phi(s_i) &=& \sum_{S \subset N \setminus \{s_i\}, T = N \setminus \{S,s_i\}}{\frac{|S|!(n-|S|-1)!}{n!}}(\vf'(\forall s_S \in S: s_S=1, s_i=1, \forall s_T \in T: s_T=0)  \nonumber\\
&-& \vf'(\forall s_S \in S: s_S=1, s_i=0,  \forall s_T \in T: s_T=0)) \label{eqn:phivec2}
\end{eqnarray}
The feature importances of variables in a baseline Shapley stands for the contribution made by each of the feature to the difference in the function between baseline and target - $\vf^{target} - \vf^{baseline}$. Hence, by the properties of Shapley values we have -
\begin{eqnarray}
&&\vf'(s_1=0, s_2=0 \hdots s_k=0) - \vf'(s_1=1, s_2=1 \hdots s_k=1) \nonumber \\
&=& \vf(P_{t}(z_1,z_2 \hdots z_{m_1}),P_{t}(z_{m_1+1} | z_{m_1 + 2} \hdots z_{m_2}) \hdots P_{t}(z_{m_j},z_{m_j+1} \hdots z_n)) \nonumber \\
&&- \vf(P_{b}(z_1,z_2 \hdots z_{m_1}),P_{b}(z_{m_1+1} | z_{m_1 + 2} \hdots z_{m_2}) \hdots P_{b}(z_{m_j},z_{m_j+1} \hdots z_n)) \nonumber \\
&=& \sum_{i=1 \hdots k}{\vphi(s_i)}
\label{eqn:vecdiffimp}
\end{eqnarray}


Eq.(\ref{eqn:phivec2}) provides a formulation of Shapley values used to compute the effect of the change in data distributions on the value of the function $f'$(equivalently $f$) defined over these distributions. The function $f$ can take any form depending on the application. In the next section, we examine a specific choice of $f$ that allows us to track the performance of a model with respect to changes in the data distribution and assign importances to specific components of the distribution.

\section{Explaining concept drift}
Concept drift refers to the phenomenon where the quality of predictions made by a model changes, in response to changes in the data. Typically, the quality of predictions made by a model deteriorates when data over which the model is evaluated is different from the data used to train the model. The change in properties of the data can either be due to a virtual drift or real drift. In this scenario, we face the challenge of separating out the factors instrumental in the change in the model performance.

We start our analysis with a general setting of supervised learning where we have input variable $X$ and an output variable $Y$. We then use a sample from their joint distribution $P_{train} (y,x)$ to learn a function $q: X \rightarrow Y$.  The hypothesis or model $q$ is often learned by minimizing the risk associated with $q$ and is defined by -
\begin{equation}
R(q) = \expect[L(q(x),y)]
\label{eqn:emprisk}
\end{equation} 
where $L$ is the loss function measuring the error between the true outcome $y$ and the prediction of the hypothesis $q(x)$. The expectation in \eq{eqn:emprisk} is measured over a data distribution $P(y,x)$. Hence, the risk $R$ measures the expected loss over predictions made by the model $q$ over a distribution $P(y,x)$. The value of risk would be different for different distributions $P(y,x)$. These distributions can be different from the distribution $P_{train}(y,x)$ which was used to train $q$. For an arbitrary distribution $P_{b}(y,x)$, \eq{eqn:emprisk} can be expanded as,
\begin{equation}
R_{baseline}(q) = \sum_{x,y}{L(q(x),y)P_{b}(y,x)}
\end{equation}
 In case of a concept drift, the distribution over the data changes to $P_{t}(y,x)$ resulting in a change in the risk over the learned model given by -
\begin{equation}
R_{target}(q) = \sum_{x,y}{L(q(x),y)P_{t}(y,x)}
\end{equation}

We can now factorise the joint distribution $P(y,x)$ into a conditional distribution $P(y|x)$ and the distribution over the input $P(x)$ such that $P(y,x) = P(y|x)P(x)$. The difference between $P_{b}(y,x)$ to $P_{t}(y,x)$ can arise due to a change in the conditional distribution and/or a change in the input distribution. The change in conditional distribution would correspond to real drift and a change in the input distribution would correspond to virtual drift. We can now rewrite risk as a function of the hypothesis, conditional distribution and the input data distribution, $R(q,P(y|x),P(x))$. Risk being a function of distribution, we can use the framework developed in Section \ref{sec:dist} to compare $R(q,P_{b}(y|x),P_{b}(x))$ to $R(q,P_{t}(y|x),P_{t}(x))$ using Shapley values. Using the Shapley framework, we can quantify the contribution made by the distributional change in the concept $P(y|x)$ and the data distribution $P(x)$, to the risk of the hypothesis $q$. Hence, using \eq{eqn:phivec2} we obtain expressions for $\phi(P(y|x))$ and $\phi(P(x))$ as -
\begin{eqnarray}
\phi(P(y|x)) &=& \frac{1}{2}\left[R(q,P_{t}(y|x),P_{b}(x)) + R(q,P_{t}(y|x),P_{t}(x)) \right. \nonumber \\
&&\left. - R(q,P_{b}(y|x),P_{b}(x)) - R(q,P_{b}(y|x),P_{t}(x))\right] \label{eqn:phicond}\\
\phi(P(x)) &=& \frac{1}{2}\left[R(q,P_{b}(y|x),P_{t}(x)) + R(q,P_{t}(y|x),P_{t}(x)) \right. \nonumber \\
&& \left. - R(q,P_{b}(y|x),P_{b}(x)) - R(q,P_{t}(y|x),P_{b}(x))\right] \label{eqn:phiinp}
\end{eqnarray}
such that,
\begin{equation}
R(q,P_{t}(y|x),P_{t}(x)) - R(q,P_{b}(y|x),P_{b}(x)) = \phi(P(y|x)) + \phi(P(x))
\end{equation}

We now illustrate the use of Shapley values to explain changes in risk, using a toy example. We construct a case where we have three features $x_1, x_2$ and $x_3$, all of them taking values from $\{0,1\}$. The dependent variable $y$ is also binary valued. The conditional distributions and the input distributions for the baseline and the target data are given in \fig{fig:scenario1} and \fig{fig:scenario2}. We assume that the hypothesis is given by the simple conjunction rule - 
\begin{equation}
q(x) = x_1 \wedge x_2 \wedge x_3
\label{eqn:hyp}
\end{equation} 
and the loss function $L$ is a mis-classification loss. We consider different scenarios of drift and analyse the feature importances provided by Shapley values in each of these cases.
\begin{itemize}
\item Real drift: In this case, we assume that the conditional distribution changes between baseline and target with the distribution over the input data remaining the same. We further assume that the ground truth relation between the input and output is the conjunction of all the three variables as given by \eq{eqn:hyp} for the baseline and turns to disjunction of the three variables - $y = x_1 \vee x_2 \vee x_3$ for the target data. The data and the conditional distributions are given in \fig{fig:scenario1}. We use these probabilities to compute the risk to be $R_{baseline} = 0$ and $R_{target} = 0.75$. We can now use \eq{eqn:phicond} and \eq{eqn:phiinp} to compute the feature importances of the change in distribution as $\phi(P(y|x)) = 0.75$ and $\phi(P(x)) = 0$. Hence, we find from the contributions that the change in risk is entirely explained by change in conditional distribution(real drift). The feature importances corroborate with our experimental design of the data.
\begin{figure}
\begin{center}
\includegraphics[width=0.9\textwidth]{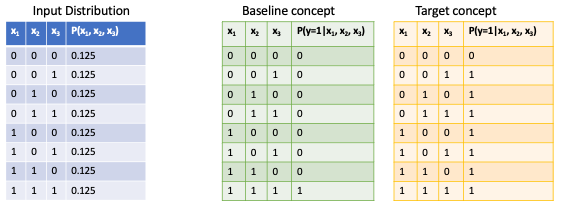}
\caption{Illustration of concept drift on toy data}
\label{fig:scenario1}
\end{center}
\end{figure}
\item Virtual drift: In this case, we assume that the underlying function mapping inputs to outputs does not change between baseline and target. The input distribution between the baseline and target changes resulting in a change in the risk. The ground truth is assumed to be a disjunction of the three variables and the shift in the input distribution is engineered in such a way that the probability of inputs that lead to erroneous predictions increase, leading to an increased risk in target compared to baseline. The details of this scenario are illustrated in \fig{fig:scenario2}. The risks corresponding to the scenario are $R_{baseline} = 0.75$ and $R_{target} = 0.78$. Corresponding feature importances can be computed using Shapley values as $\phi(P(y|x)) = 0$ and $\phi(P(x)) = 0.03$. These values for feature importances confirm with the data used in this scenario thus validating our approach.
\begin{figure}
\begin{center}
\includegraphics[width=0.9\textwidth]{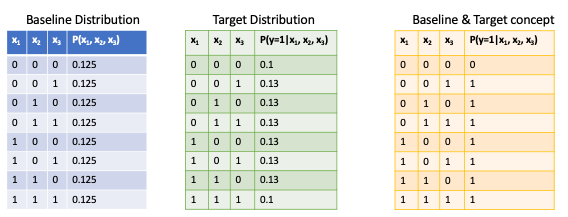}
\caption{Illustration of concept drift on toy data}
\label{fig:scenario2}
\end{center}
\end{figure}
\end{itemize}

In real world, drift is usually a combination of both virtual and real drifts. In such circumstances, this framework of using Shapley values can help quantify the contributions of virtual and real drifts to the overall drift. We combine the scenarios given in \fig{fig:scenario1} and \fig{fig:scenario2} to find the difference in risk with the combined changes in data. In this combination of scenarios, we obtain the risks to be $R_{target}  =  0.78$ and $R_{baseline} = 0$. The feature importances attributed to real and virtual drifts are given by $\phi(P(y|x)) = 0.765$ and $\phi(P(x)) = 0.015$ respectively.

Hence, with these examples we have illustrated the use of Shapley values in attributing the change in the risk of a model to the changes in the underlying real and virtual drifts in data distribution. The framework that we have developed works theoretically but requires some approximations for it to work in practical applications. In the next section, we sketch the algorithm for explaining the drivers behind drift in practical situations. We call this algorithm Distribution Baseline Shap(\ouralgo).
\section{ \ouralgo for explaining drift}
In this section, we detail the approximations we make to the Shapley values based analysis of concept drift. This helps in applying this framework to real world data.

\begin{itemize}
\item Binning: The framework we have developed works for both continuous as well as discrete distributions of data. However, with continuous distributions we would have to fit a parametric continuous distribution to the input variable and work with it to compute Shapley values analytically. The other alternative would be to bin the values of continuous variables into different buckets and use those as categories for the variable.
\item Empirical risk: The definition of risk $R$ given in \eq{eqn:emprisk} generally cannot be computed because the data distribution $P(y,x)$ is unknown. The usual approximation made in machine learning literature is to approximate this risk using the empirical estimate of the distribution from the test data. Hence, the empirical risk is given by -
\begin{equation}
\hat{R}(q) = \sum_{x,y}{L(q(x),y)\hat{P}(y,x)}
\end{equation}
where $\hat{P}$ is the empirical estimate of the data distribution and $\hat{R}$ the corresponding empirical risk estimate. We compute the empirical estimates of distributions $P(y|x)$ and $P(x)$ as probability tables over discretized variables. These distributions are estimated from the data corresponding to baseline and target populations. 
\item Shap: In many applications, we might be interested in measuring the importance of individual variables to the drift. This can be accomplished by factorising the input distribution based on the dependency structure of the inputs. The simplest of this factorization could be to treat the individual input variables as independent of each other and factorise the joint distribution as a product of the marginal distribution of individual features. However, with increasing number of features, computation of exact Shapley values would become intractable. To avoid intractability, we use Baseline SHAP(\cite{shap}) to compute approximate Shapley values.
\end{itemize}

The final algorithm using the approximations listed above is given in Algorithm \ref{algo}.

\begin{algorithm}
\SetAlgoLined
\KwData{Hypothesis $q$,Baseline population data, Target population data}
\KwResult{Importances of real drift and virtual drift}
\If{continuous feature}
	{Bin values into finite bins for features with continuous values\;}
{Estimate $P_{b}(y|x)$ and $P_{b}(x)$ from the data\;}
{Estimate $P_{t}(y|x)$ and $P_{t}(x)$ from the data\;}
{Restructure the function into $f'$ using surrogate features as given in \eq{eqn:s}\;}
{Define the risk function $R$ based on a selected loss, hypothesis $q$ and the distributions\;}
{Compute the approximate Shapley values using risk $R$ and output the importances\;}
\caption{Description of \ouralgo algorithm}
\label{algo}
\end{algorithm}
\section{Evaluation}
In this section, we perform various experiments to evaluate the efficacy of \ouralgo. We use benchmark datasets commonly used in concept drift studies(\cite{survey}) to measure the efficacy. The datasets we have used is a collection of synthetically generated data and data from real world. In the synthetically generated datasets, the data is generated by a known process and the concept drift is induced across different segments of the data. Hence, the ground truth regarding the drift is known and can be used to evaluate the effectiveness of \ouralgo. For each of the synthetic datasets, there are two different regimes - one before the drift and one after the drift. We use the former as the baseline and the latter as target population. We apply \ouralgo to compare the performance of the model between the two regimes and record the relative importances attributed to real drift and virtual drift.  In these experiments, as with all explainability tools it is difficult to verify the magnitudes of importance provided by the tool. Hence, we evaluate the effectiveness of the algorithm by making sure that the algorithm is able to match the relative importances between real and virtual drift as present in the ground truth. The results of the evaluations are provided in Table \ref{tab:synth}. For each dataset, the data was split into a baseline regime and a target regime. A simple decision tree with depth adapted to each dataset was trained using the baseline data. This forms our hypothesis $q$ referred in Algorithm \ref{algo}. The probability distributions are estimated from the data by computing the frequency of occurrence of various combinations of values. The risk was evaluated using the misclassification loss. In all the experiments, we make an assumption that the input features are independent and hence the joint distribution of input variables factorise into marginal distributions over individual variables. This helps us measure the importances of distributions over individual features. 

In Table \ref{tab:synth}, we evaluate \ouralgo over two different scenarios. In the first scenario, there is a real drift introduced in the data between baseline and target. This is the same as the original design of the dataset and has been used in many of the papers to detect concept drift. As we can see from the columns under {\em Change in conditional dist.}, the real drift dominates over the virtual drift for all the datasets as expected. Two of the features with the highest importances are reported in the virtual drift column.  In the second scenario, the underlying concept is held constant but the input distribution of one particular feature is altered by multiplying the values of the target data by random numbers generated from a uniform distribution. This ensures that the range of values taken by the feature in the target data is different from the one used to train the hypothesis. As can be seen from columns marked {\em Change in input dist.}, the contribution of the virtual drift has increased substantially from that of its contribution in the first scenario. For instance, in the STAGGER dataset the importance of the first feature has increased from $9.06\times10^{-3}$ in the first scenario to $0.01$ in the second scenario. Although the underlying concept was not changed between baseline and target, there is noise added to the underlying concept. Hence the contribution from real drift is not zero.The features that were identified to have the largest contribution also matches the ones whose distributions were altered. Hence, these experiments demonstrate the efficacy of \ouralgo in quantifying the contribution of different components of the data, to the drift in model performance.

In the next set of experiments, we evaluate \ouralgo over real world data that are often used as benchmark, in research related to concept drift. The results over the real world datasets are given in Table \ref{tab:real}. The experimental setting is similar to the one on synthetic data with the hypothesis being a decision tree of depth adapted to the dataset and the loss function being the misclassification error. However, unlike the synthetic data, we do not know the partition of the data where the concept drift has occurred. It could also happen in these cases that the concept drift is gradual. Hence, for our experiments, we have split the data into equal halves with one being treated as baseline and the other as target. In Electricity dataset where there is a temporal ordering to the data, we sort the data according to its timeline and then choose the first half for baseline and second half for target. From the results in Table \ref{tab:real} we find that the contribution from real drift and virtual drift can vary between datasets and hypothesis. Hence, using \ouralgo we can better understand the root cause of drift in real world models and use this to diagnose and rectify the model.
\begin{table}
\begin{center}
\begin{tabular}{|p{0.6in}|p{0.5in}|p{0.5in}|p{0.4in}|p{0.4in}|p{0.4in}|p{0.4in}|p{0.4in}|p{0.4in}|}
\hline
&&&\multicolumn{3}{c|}{Change in conditional dist.} & \multicolumn{3}{c|}{Change in input dist.}\\
\hline
Dataset&No. of features&No. of classes&Real drift&\multicolumn{2}{c|}{Virtual drift}&Real drift&\multicolumn{2}{c|}{Virtual drift}\\
\hline
STAGGER&3&2&\bf{0.58}&9.06e-3&-7.1e-3&-0.08&\bf{0.01}&7.46e-3\\
\hline
SEA&3&2&\bf{0.30}&-2.59e-4&-6.02e-5&3.55e-4&\bf{-2.27e-2}&2.02e-4\\
\hline
Sine&2&2&\bf{0.22}&3.8e-5&-2.27e-6&-0.07&\bf{-0.03}&1.88e-4\\
\hline
Circle&2&2&\bf{0.16}&-1.4e-2&8.67e-6&-0.02&\bf{-0.10}&-4.39e-6\\
\hline
\end{tabular}
\caption{Contributions of real and virtual drift on synthetic datasets}
\label{tab:synth}
\end{center}
\end{table}

\begin{table}
\begin{center}
\begin{tabular}{|p{0.5in}|p{0.6in}|p{0.6in}|p{0.4in}|p{1.2in}|p{1.2in}|}
\hline
Dataset&No. of features&No. of classes&Real drift&\multicolumn{2}{c|}{Virtual drift}\\
\hline
Electricity&7&2&0.05&transfer: -0.06&vicdemand:-0.064\\
\hline
Airlines&7&2&0.01&Airline:3.8e-4&Length:2e-4\\
\hline
Covertype&54&7&1.45e-3&Horizontal dist \newline to roadways: 1.59e-3&Horizontal dist \newline to fire points: 1.29e-3\\
\hline
\end{tabular}
\caption{Contributions of real and virtual drift towards the drifts. The values in the virtual drift column specifies the feature name and its corresponding Shapley value.}
\label{tab:real}
\end{center}
\end{table}

In the final experiment, we measure the computational efficiency of \ouralgo. We use another synthetic dataset - RBF to measure the computational requirement of \ouralgo. The computational complexity of the algorithm grows with the number of features, since the number of possible combinations to compute Shapley increases with the number of features. However, as seen from Table \ref{tab:comp}, \ouralgo scales well with the increase in feature, primarily due to the approximations listed in Algorithm \ref{algo} to compute Shapley values.
\begin{table}[h]
\begin{center}
\begin{tabular}{| *{5}{c|}}
\hline
No. of features&Real drift&\multicolumn{2}{c|}{Virtual drift}&Compute time(in sec)\\
\hline
10&0.18&0.07&0.05&64.76(1.4)\\
\hline
15&0.03&-0.02&0.01&192.66(8.2)\\
\hline
20&7.3e-4&9.1e-4&2.6e-4&695.5(94)\\
\hline
\end{tabular}
\caption{Provides the time taken for evaluating the Shapley values with increasing number of features. The numbers in the parenthesis corresponds to the standard deviation over 3 runs.}
\label{tab:comp}
\end{center}
\end{table}

\section{Conclusion}
In this paper, we have introduced a novel explainability framework that can be used to compare different populations and understand the driving factors behind the change in the quality of predictions between these populations. In order to enable this, we have extended Shapley values to be able to utilise distributions functions as features and compute the effect of the changes in distributions to the overall change in the accuracy metrics of the model. We have also provided empirical evidence to support the framework through benchmarks on synthetic and real world dataset with drifts.

The ideas presented in this paper, though simple, is novel and impactful. Use of Shapley values to explain the change in model behaviour with drift in data is novel and can be used in a number of applications. The mathematical framework developed in this framework is generic and can be applied to different forms of model, loss function and data distributions. Apart from using it in understanding concept drift, it can be used to aid decisions on when to retire or retrain a model in response to change in its behaviour, thus being useful in a number of applications.
\section*{Broader Impact}
This research can be useful to the larger society when we want to explain the differences between two subpopulations. As had been demonstrated in the paper, it can be used to understand and quantify the extent of unfairness contributed by different factors of the data while evaluating a model. \ouralgo can help diagnose the shortcomings of black box models and help to make them more adaptable to changing and unforeseen events like COVID or global recession. The ideas proposed in this paper does not affect any society or group of individual in a negative way. 
\section*{Limitation}
The limitation of \ouralgo is the approximations we make to make it practically feasible. If these approximations are violated in the data it can lead to incorrect values of importance. The other limitation which is shared by all explainability methods is the absence of any robust methodology to measure the accuracy of the importance values produced by algorithms like \ouralgo except for synthetically generated datasets.
\bibliographystyle{ACM-Reference-Format}
\bibliography{paper}


\begin{thebibliography}{13}


\ifx \showCODEN    \undefined \def \showCODEN     #1{\unskip}     \fi
\ifx \showDOI      \undefined \def \showDOI       #1{#1}\fi
\ifx \showISBNx    \undefined \def \showISBNx     #1{\unskip}     \fi
\ifx \showISBNxiii \undefined \def \showISBNxiii  #1{\unskip}     \fi
\ifx \showISSN     \undefined \def \showISSN      #1{\unskip}     \fi
\ifx \showLCCN     \undefined \def \showLCCN      #1{\unskip}     \fi
\ifx \shownote     \undefined \def \shownote      #1{#1}          \fi
\ifx \showarticletitle \undefined \def \showarticletitle #1{#1}   \fi
\ifx \showURL      \undefined \def \showURL       {\relax}        \fi
\providecommand\bibfield[2]{#2}
\providecommand\bibinfo[2]{#2}
\providecommand\natexlab[1]{#1}
\providecommand\showeprint[2][]{arXiv:#2}

\bibitem[\protect\citeauthoryear{Adams, van Zelst, Quack, Hausmann, van~der
  Aalst, and Rose}{Adams et~al\mbox{.}}{2021}]%
        {processmining}
\bibfield{author}{\bibinfo{person}{Jan~Niklas Adams},
  \bibinfo{person}{Sebastiaan~J. van Zelst}, \bibinfo{person}{Lara Quack},
  \bibinfo{person}{Kathrin Hausmann}, \bibinfo{person}{Wil M.~P. van~der
  Aalst}, {and} \bibinfo{person}{Thomas Rose}.}
  \bibinfo{year}{2021}\natexlab{}.
\newblock \showarticletitle{A Framework for Explainable Concept Drift Detection
  in Process Mining}. In \bibinfo{booktitle}{\emph{Business Process
  Management}}, \bibfield{editor}{\bibinfo{person}{Artem Polyvyanyy},
  \bibinfo{person}{Moe~Thandar Wynn}, \bibinfo{person}{Amy Van~Looy}, {and}
  \bibinfo{person}{Manfred Reichert}} (Eds.). \bibinfo{publisher}{Springer
  International Publishing}, \bibinfo{address}{Cham},
  \bibinfo{pages}{400--416}.
\newblock
\showISBNx{978-3-030-85469-0}


\bibitem[\protect\citeauthoryear{Alippi and Roveri}{Alippi and Roveri}{2008}]%
        {detect1}
\bibfield{author}{\bibinfo{person}{Cesare Alippi} {and} \bibinfo{person}{Manuel
  Roveri}.} \bibinfo{year}{2008}\natexlab{}.
\newblock \showarticletitle{Just-in-Time Adaptive Classifiers—Part I:
  Detecting Nonstationary Changes}.
\newblock \bibinfo{journal}{\emph{IEEE Transactions on Neural Networks}}
  \bibinfo{volume}{19}, \bibinfo{number}{7} (\bibinfo{year}{2008}),
  \bibinfo{pages}{1145--1153}.
\newblock
\urldef\tempurl%
\url{https://doi.org/10.1109/TNN.2008.2000082}
\showDOI{\tempurl}


\bibitem[\protect\citeauthoryear{Dasu, Krishnan, Venkatasubramanian, and
  Yi}{Dasu et~al\mbox{.}}{2006}]%
        {detect2}
\bibfield{author}{\bibinfo{person}{Tamraparni Dasu}, \bibinfo{person}{Shankar
  Krishnan}, \bibinfo{person}{Suresh Venkatasubramanian}, {and}
  \bibinfo{person}{Ke Yi}.} \bibinfo{year}{2006}\natexlab{}.
\newblock \showarticletitle{An Information-Theoretic Approach to Detecting
  Changes in MultiDimensional Data Streams}.
\newblock \bibinfo{journal}{\emph{Interfaces}} (\bibinfo{date}{01}
  \bibinfo{year}{2006}).
\newblock


\bibitem[\protect\citeauthoryear{Gama, \v{Z}liobaitundefined, Bifet,
  Pechenizkiy, and Bouchachia}{Gama et~al\mbox{.}}{2014}]%
        {gama}
\bibfield{author}{\bibinfo{person}{Jo\~{a}o Gama},
  \bibinfo{person}{Indrundefined \v{Z}liobaitundefined},
  \bibinfo{person}{Albert Bifet}, \bibinfo{person}{Mykola Pechenizkiy}, {and}
  \bibinfo{person}{Abdelhamid Bouchachia}.} \bibinfo{year}{2014}\natexlab{}.
\newblock \showarticletitle{A Survey on Concept Drift Adaptation}.
\newblock \bibinfo{journal}{\emph{ACM Comput. Surv.}} \bibinfo{volume}{46},
  \bibinfo{number}{4}, Article \bibinfo{articleno}{44} (\bibinfo{date}{mar}
  \bibinfo{year}{2014}), \bibinfo{numpages}{37}~pages.
\newblock
\showISSN{0360-0300}
\urldef\tempurl%
\url{https://doi.org/10.1145/2523813}
\showDOI{\tempurl}


\bibitem[\protect\citeauthoryear{Gao, Fan, Han, and Yu}{Gao
  et~al\mbox{.}}{2007}]%
        {gao}
\bibfield{author}{\bibinfo{person}{Jing Gao}, \bibinfo{person}{Wei Fan},
  \bibinfo{person}{Jiawei Han}, {and} \bibinfo{person}{Philip Yu}.}
  \bibinfo{year}{2007}\natexlab{}.
\newblock \showarticletitle{A General Framework for Mining Concept-Drifting
  Data Streams with Skewed Distributions}.
\newblock
\urldef\tempurl%
\url{https://doi.org/10.1137/1.9781611972771.1}
\showDOI{\tempurl}


\bibitem[\protect\citeauthoryear{Krawczyk, Minku, Gama, Stefanowski, and
  Woźniak}{Krawczyk et~al\mbox{.}}{2017}]%
        {adapt1}
\bibfield{author}{\bibinfo{person}{Bartosz Krawczyk},
  \bibinfo{person}{Leandro~L. Minku}, \bibinfo{person}{João Gama},
  \bibinfo{person}{Jerzy Stefanowski}, {and} \bibinfo{person}{Michał
  Woźniak}.} \bibinfo{year}{2017}\natexlab{}.
\newblock \showarticletitle{Ensemble learning for data stream analysis: A
  survey}.
\newblock \bibinfo{journal}{\emph{Information Fusion}}  \bibinfo{volume}{37}
  (\bibinfo{year}{2017}), \bibinfo{pages}{132--156}.
\newblock
\showISSN{1566-2535}
\urldef\tempurl%
\url{https://doi.org/10.1016/j.inffus.2017.02.004}
\showDOI{\tempurl}


\bibitem[\protect\citeauthoryear{Lazarescu, Venkatesh, and Bui}{Lazarescu
  et~al\mbox{.}}{2004}]%
        {laz}
\bibfield{author}{\bibinfo{person}{Mihai Lazarescu}, \bibinfo{person}{Svetha
  Venkatesh}, {and} \bibinfo{person}{Hung Bui}.}
  \bibinfo{year}{2004}\natexlab{}.
\newblock \showarticletitle{Using multiple windows to track concept drift}.
\newblock \bibinfo{journal}{\emph{Intell. Data Anal.}}  \bibinfo{volume}{8}
  (\bibinfo{date}{03} \bibinfo{year}{2004}), \bibinfo{pages}{29--59}.
\newblock
\urldef\tempurl%
\url{https://doi.org/10.3233/IDA-2004-8103}
\showDOI{\tempurl}


\bibitem[\protect\citeauthoryear{Lu, Liu, Dong, Gu, Gama, and Zhang}{Lu
  et~al\mbox{.}}{2018}]%
        {survey}
\bibfield{author}{\bibinfo{person}{Jie Lu}, \bibinfo{person}{Anjin Liu},
  \bibinfo{person}{Fan Dong}, \bibinfo{person}{Feng Gu}, \bibinfo{person}{Joao
  Gama}, {and} \bibinfo{person}{Guangquan Zhang}.}
  \bibinfo{year}{2018}\natexlab{}.
\newblock \showarticletitle{Learning under Concept Drift: A Review}.
\newblock \bibinfo{journal}{\emph{{IEEE} Transactions on Knowledge and Data
  Engineering}} (\bibinfo{year}{2018}), \bibinfo{pages}{1--1}.
\newblock
\urldef\tempurl%
\url{https://doi.org/10.1109/tkde.2018.2876857}
\showDOI{\tempurl}


\bibitem[\protect\citeauthoryear{Lundberg and Lee}{Lundberg and Lee}{2017}]%
        {shap}
\bibfield{author}{\bibinfo{person}{Scott~M Lundberg} {and}
  \bibinfo{person}{Su-In Lee}.} \bibinfo{year}{2017}\natexlab{}.
\newblock \showarticletitle{A Unified Approach to Interpreting Model
  Predictions}. In \bibinfo{booktitle}{\emph{Advances in Neural Information
  Processing Systems}}, \bibfield{editor}{\bibinfo{person}{I.~Guyon},
  \bibinfo{person}{U.~Von Luxburg}, \bibinfo{person}{S.~Bengio},
  \bibinfo{person}{H.~Wallach}, \bibinfo{person}{R.~Fergus},
  \bibinfo{person}{S.~Vishwanathan}, {and} \bibinfo{person}{R.~Garnett}}
  (Eds.), Vol.~\bibinfo{volume}{30}. \bibinfo{publisher}{Curran Associates,
  Inc.}
\newblock
\urldef\tempurl%
\url{https://proceedings.neurips.cc/paper_files/paper/2017/file/8a20a8621978632d76c43dfd28b67767-Paper.pdf}
\showURL{%
\tempurl}


\bibitem[\protect\citeauthoryear{Ramírez-Gallego, Krawczyk, García, Woźniak,
  and Herrera}{Ramírez-Gallego et~al\mbox{.}}{2017}]%
        {adapt2}
\bibfield{author}{\bibinfo{person}{Sergio Ramírez-Gallego},
  \bibinfo{person}{Bartosz Krawczyk}, \bibinfo{person}{Salvador García},
  \bibinfo{person}{Michał Woźniak}, {and} \bibinfo{person}{Francisco
  Herrera}.} \bibinfo{year}{2017}\natexlab{}.
\newblock \showarticletitle{A survey on data preprocessing for data stream
  mining: Current status and future directions}.
\newblock \bibinfo{journal}{\emph{Neurocomputing}}  \bibinfo{volume}{239}
  (\bibinfo{year}{2017}), \bibinfo{pages}{39--57}.
\newblock
\showISSN{0925-2312}
\urldef\tempurl%
\url{https://doi.org/10.1016/j.neucom.2017.01.078}
\showDOI{\tempurl}


\bibitem[\protect\citeauthoryear{Salganicoff}{Salganicoff}{1993}]%
        {sal}
\bibfield{author}{\bibinfo{person}{Marcos Salganicoff}.}
  \bibinfo{year}{1993}\natexlab{}.
\newblock \showarticletitle{Density-Adaptive Learning and Forgetting}. In
  \bibinfo{booktitle}{\emph{Proceedings of the Tenth International Conference
  on International Conference on Machine Learning}} (Amherst, MA, USA)
  \emph{(\bibinfo{series}{ICML'93})}. \bibinfo{publisher}{Morgan Kaufmann
  Publishers Inc.}, \bibinfo{address}{San Francisco, CA, USA},
  \bibinfo{pages}{276–283}.
\newblock
\showISBNx{1558603077}


\bibitem[\protect\citeauthoryear{Sundararajan and Najmi}{Sundararajan and
  Najmi}{2020}]%
        {bshap}
\bibfield{author}{\bibinfo{person}{Mukund Sundararajan} {and}
  \bibinfo{person}{Amir Najmi}.} \bibinfo{year}{2020}\natexlab{}.
\newblock \showarticletitle{The Many Shapley Values for Model Explanation}. In
  \bibinfo{booktitle}{\emph{Proceedings of the 37th International Conference on
  Machine Learning}} \emph{(\bibinfo{series}{Proceedings of Machine Learning
  Research}, Vol.~\bibinfo{volume}{119})},
  \bibfield{editor}{\bibinfo{person}{Hal~Daumé III} {and}
  \bibinfo{person}{Aarti Singh}} (Eds.). \bibinfo{publisher}{PMLR},
  \bibinfo{pages}{9269--9278}.
\newblock
\urldef\tempurl%
\url{https://proceedings.mlr.press/v119/sundararajan20b.html}
\showURL{%
\tempurl}


\bibitem[\protect\citeauthoryear{Widmer and Kubat}{Widmer and Kubat}{1996}]%
        {stagger}
\bibfield{author}{\bibinfo{person}{Gerhard Widmer} {and}
  \bibinfo{person}{Miroslav Kubat}.} \bibinfo{year}{1996}\natexlab{}.
\newblock \showarticletitle{Learning in the presence of concept drift and
  hidden contexts}.
\newblock \bibinfo{journal}{\emph{Machine Learning}} \bibinfo{volume}{23},
  \bibinfo{number}{1} (\bibinfo{year}{1996}), \bibinfo{pages}{69--101}.
\newblock
\showISBNx{1573-0565}
\urldef\tempurl%
\url{https://doi.org/10.1007/BF00116900}
\showDOI{\tempurl}


\end{thebibliography}

\end{document}